# Scaling Intelligent Agents in Combat Simulations for Wargaming

**Scotty Black**
Naval Postgraduate School
Monterey, California
scotty.black@nps.edu

**Christian Darken**
Naval Postgraduate School
Monterey, California
cjdarken@nps.edu

## ABSTRACT

Remaining competitive in future conflicts with technologically-advanced competitors requires us to accelerate our research and development in artificial intelligence (AI) for wargaming. More importantly, leveraging machine learning for intelligent combat behavior development will be key to one day achieving superhuman performance in this domain—elevating the quality and accelerating the speed of our decisions in future wars. Although deep reinforcement learning (RL) continues to show promising results in intelligent agent behavior development in games, it has yet to perform at or above the human level in the long-horizon, complex tasks typically found in combat modeling and simulation. Capitalizing on the proven potential of RL and recent successes of hierarchical reinforcement learning (HRL), our research is investigating and extending the use of HRL to create intelligent agents capable of performing effectively in these large and complex simulation environments. Our ultimate goal is to develop an agent capable of superhuman performance that could then serve as an AI advisor to military planners and decision-makers. This papers covers our ongoing approach and the first three of our five research areas aimed at managing the exponential growth of computations that have thus far limited the use of AI in combat simulations: (1) developing an HRL training framework and agent architecture for combat units; (2) developing a multi-model framework for agent decision-making; (3) developing dimension-invariant observation abstractions of the state space to manage the exponential growth of computations; (4) developing an intrinsic rewards engine to enable long-term planning; and (5) implementing this framework into a higher-fidelity combat simulation. This research will further the ongoing Department of Defense's research interest in scaling AI to deal with large and complex military scenarios in support of wargaming for concept development, education, and analysis.

## ABOUT THE AUTHORS

**Scotty Black** is a Lieutenant Colonel in the U.S. Marine Corps assigned to the Naval Postgraduate School as part of the Marine Corps Technical PhD Program. His primary specialty is as an F/A-18 Weapons Systems Officer with over 18 years of Marine Corps experience. LtCol Black is currently a PhD Candidate conducting research leveraging hierarchical reinforcement learning to scale artificial intelligence to deal with the often large and complex state spaces inherent in combat modeling and simulation for wargaming. LtCol Black's experience includes nearly 2,000 flight hours in the F/A-18, graduate of the Weapons and Tactics Instructor (WTI) Course, multiple combat deployments, leading science and technology initiative for Marine Corps training and education as a Modeling and Simulation Officer, a DARPA Service Chiefs Fellowship, and research fellowships at the Naval Information Warfare Center Pacific and the former Space and Naval Warfare Systems Center Pacific.

**Christian Darken** is an Associate Professor in the Department of Computer Science at the Naval Postgraduate School, where he is also a member of the MOVES (Modeling, Virtual Environments and Simulation) faculty. He has more than 35 years of machine learning research experience and has been conducting teaching and research on human behavior models for simulations for over twenty years. His background includes technical program management for Siemens Corporation, and serving as program and general chair for the AAAI-sponsored AIIDE conference.

# Scaling Intelligent Agents in Combat Simulations for Wargaming

**Scotty Black**

**Naval Postgraduate School**

**Monterey, California**

**scotty.black@nps.edu**

**Christian Darken**

**Naval Postgraduate School**

**Monterey, California**

**cjdarken@nps.edu**

## INTRODUCTION

Recent advances in artificial intelligence (AI) technologies, such as OpenAI's ChatGPT, have once again exemplified the transformative potential AI can have on reshaping various industries. Just as generative pre-trained transformers (GPT) models have radically redefined our understanding of the tremendous power of AI, so can other AI methodologies contribute to the development of game-changing tools for defense sectors that have thus far proven too complex for AI to effectively address.

One such area where AI can have a transformative impact is in the domain of combat modeling and simulation in support of wargaming. Unfortunately, however, much like the history of military wargaming goes back centuries, so do most of the tools and techniques still used today to conduct modern wargaming. While there is absolutely still a role for traditional wargaming artifacts such as physical game boards, cards, and dice, there is increasing pressure to bring wargaming into the 21st Century (Berger, 2020, 2022; Defense Science Board, 2021; Deputy Secretary of Defense, 2015; United States Government Accountability Office, 2023) and leverage modern technological advances, such as AI (Davis & Bracken, 2022) "to evolve the current paradigm of wargaming—both in terms of technology and methodology" (Wong et al., 2019).

Although the U.S. has enjoyed military superiority in most domains, the democratization of machine learning (ML) has begun to provide our competitors and other state actors with innumerable opportunities for disruption (Zhang et al., 2020). Therefore, more than ever before, it becomes imperative that we aggressively invest in research and development to build a solid ground-level understanding of the strengths and weaknesses of AI (Schmidt et al., 2021) and how it can be used for designing, planning, executing, and analyzing wargames for all purposes. Only then can the Department of Defense (DOD) be better prepared to deal with strategic surprise and disruption (Zhang et al., 2020).

Wargaming and military planning, however, are significantly different than the traditional domains that have successfully leveraged AI thus far—such as image classification and logistics-related optimization problems. Because of the complexity of warfare, mission analysis and planning generally require intuition and mental heuristics to be applied early-on to limit the size of the search problem (Zhang et al., 2020). While heuristics do allow us to find acceptable solutions more easily, these solutions are typically not scalable or reliable enough to evaluate the vast number of contingencies that might arise. Furthermore, intuition can also be insufficient in addressing highly complex problems, such as those involving high-dimensional spaces with many different players and complex weapons and sensor interactions (Zhang et al., 2020)—yet these complexities are the very characteristics that may define the future of warfare (Narayanan et al., 2021).

While we do not envision AI replacing human judgment or decision-making in the foreseeable future, we do contend that AI—when incorporated into decision aids—offers a chance to speed up our decision-making process and provide new insights. In fact, failing to leverage the power of AI potentially incurs substantial risk as we venture deeper into multi-domain operations (Narayanan et al., 2021). Ultimately, by leveraging superhuman intelligent agents as the foundation for decision-support tools for human decision-makers, we expect to be able to reach a decision advantage over our adversaries—accelerating the speed and elevating the quality of our decisions in future wars. Remaining competitive in future conflicts with technologically-advanced competitors, therefore, requires us to accelerate our research and development of AI for wargaming. More importantly, leveraging ML for intelligent combat behavior development will be key to one day achieving superhuman performance in this domain.

This paper presents our research approach to scaling AI to deal with the complex and intricate state spaces characteristic of combat modeling and simulation for wargaming. While our research is still ongoing and incomplete, in this paper we present our overall approach, preliminary results, and way forward.

## BACKGROUND

### Reinforcement Learning

Reinforcement learning (RL) is a type of ML where an agent learns to make decisions simply by interacting with its environment. To learn the correct actions, the agent performs an action, receives feedback in the form of rewards or punishments, and aims to maximize the total reward over time. This trial-and-error process allows the agent to learn the best actions for different states based on the feedback it receives, which becomes its policy. This policy is essentially the strategy that the agent employs to make decisions.

More formally put, a reinforcement learning problem, shown in Figure 1, typically consists of a decision-maker, referred to as an *agent*, and an *environment*, represented by *states* $s \in S$. The *agent* can take *actions* $a_t$ as a function of the current *state* $s_t$ such that $a_t \in A(s_t)$. After choosing an *action* at time $t$ $(a_t)$, the agent first receives a *reward* $r_{t+1}$ and finds itself in a new *state* $s_{t+1}$. The *action* $a_t$ comes from a strategy called a *policy* $\pi$. This *policy* $\pi$ is a mapping from *states* $s \in S$ to a probability of selecting each possible *action* $\pi(s,a)$. As the *agent* interacts with the *environment*, it will learn the optimal *policy* that maximizes its *reward* in the long run.

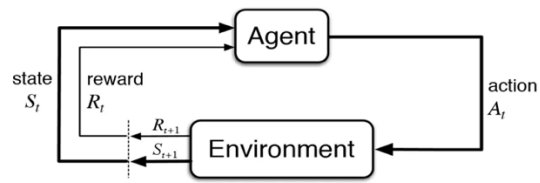

**Figure 1. The Reinforcement Learning Problem**

There exists much research employing RL to create intelligent agents capable of gameplaying. A select few we leverage for our approach include RL research exploring Atari 2600 games (Mnih et al., 2015; Van Hasselt et al., 2016), Go (Holcomb et al., 2018), Chess (Silver et al., 2017), Shogi (Silver et al., 2017), Dota 2 (Berner et al., 2019), StarCraft II (Vinyals et al., 2019), and Atlatl (Allen, 2022; Boron, 2020; Cannon & Goericke, 2020). Nevertheless, despite RL having achieved human, expert, or even superhuman-level play in some of these games, to date, no AI agent has been shown to generally outperform humans in the complex domain of wargaming.

### Hierarchical Reinforcement Learning

To effectively employ AI in combat simulations in support of wargaming, RL agents still need to be able to appropriately scale to meet the requirements of large wargames involving large state spaces and hundreds, or even thousands, of entities. However, as the number and different types of entities increase in a wargame, so does the amount of information (i.e., action and state spaces)—quickly becoming an intractable problem.

Using a hierarchical approach to deal with this exponential problem seems promising in that hierarchical decomposition is a natural way humans already break down complex tasks and re-use old or related skills (Frans et al., 2017). Using a hierarchical decomposition might facilitate the generation of complex behaviors by attempting to solve the problem at multiple levels of abstraction (Russell et al., 2010). In their chapter *Behavioral Hierarchy: Exploration and Representation*, Barto et al. discuss how behavioral modules can provide reusable building blocks that can be composed hierarchically to generate an extensive range of behaviors (Baldassarre & Mirolli, 2013). Traditional RL alone, on the other hand, must master hundreds or even thousands of small tasks independently and from scratch which, though possible for relatively simple games, is computationally very expensive and often intractable for more complex games. While there is a large body of research seeking to improve sample efficiency in RL, in the absence of prior knowledge, there currently exists a limit to learning speed (Frans et al., 2017). Therefore, RL on its own will likely fall short of scaling to be useful in more complex applications such as wargaming.

Hierarchical reinforcement learning (HRL), on the other hand, shows increasing promise to help address what has been referred to as Bellman's "curse of dimensionality" in that HRL "decomposes a reinforcement learning problem into a hierarchy of subproblems or subtasks such that higher-level parent-tasks invoke lower-level child-tasks as if they were primitive actions" (Sammut & Webb, 2010). This decomposition may itself have multiple levels of hierarchy, of which some or all could be RL problems themselves. By decomposing the problem using hierarchical models, we may be able to solve more complex problems by reducing the computational complexity given that the overall problem can be appropriately represented more compactly, and their subtasks be reused or learned independently. Of note, however, given the constraints of the hierarchy, although the solution to an HRL problem may be optimal, there is no guarantee that the decomposed solution is an optimal solution to the original problem (Sammut & Webb, 2010).

Examples of insights we intend to incorporate from the HRL literature include: leveraging the high-level networks for *task decisions* and the lower-level networks for task-associated behaviors or *primitives* (S. Li et al., 2022; Morimoto & Doya, 1998; Pope et al., 2021; Vezhnevets et al., 2017; Wang et al., 2021); incorporating variations of a *policy-selector network* (Pope et al., 2021); developing *active hierarchical exploration strategies* (S. Li et al., 2022); extending the concept of a *local observation space* (Rood, 2022); applying principles of *reward hiding* and *information hiding* (Dayan & Hinton, 1992); training each level of the hierarchy *independently* (Pope et al., 2021; Vezhnevets et al., 2017); and adding *intrinsic rewards* (Vezhnevets et al., 2017).

**Atlatl Combat Simulation Environment**

We use the Atlatl Combat Simulation environment (Darken, 2022) to develop, implement, and experiment with our research approach. Atlatl is a simple but effective combat model that was developed at the Naval Postgraduate School's Modeling, Virtual Environments, and Simulation (MOVES) Institute. It includes an underlying combat model that is deterministic and purposefully simplistic, as well as the surrounding Gymnasium (Farama Foundation, 2023) infrastructure that supports rapid AI experimentation. This type of basic environment allows researchers to build experience applying cutting-edge AI to operational and tactical problems. It allows for quick learning of the different AI algorithms' strengths and weaknesses, enabling researchers to more quickly answer research questions relating to applying different AI approaches to combat modeling and simulation.

Currently, performance is scored based on kills, losses, and holding urban areas (as a surrogate for all seize and hold operations), though these metrics can easily be changed as desired. The environment also contains hooks that enable interfacing with standard RL codebase and algorithms, such as Stable Baselines 3.

Through a web browser interface, Atlatl allows a human player to play against the AI, however, the simulation can also run headless with an AI playing against another AI. Additionally, a browser-based replay capability allows for replays of AI versus AI engagement. An example of a simple scenario on an Atlatl game board is shown in Figure 2. Units are represented visually by their respective military operational terms and graphics, and terrain is represented visually in colors (e.g., water is blue, rough terrain is brown, urban is gray).

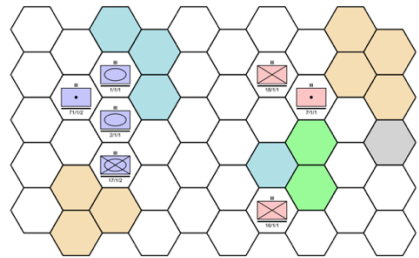

**Figure 2. Atlatl Gameboard Example**

**RESEARCH PLAN**

Capitalizing on the demonstrated potential of RL and recent successes of HRL, our research intends to further the ability to scale machine learning to develop intelligent agent behaviors for use in the large and complex scenarios typically found in combat modeling and simulation. To accomplish this, we intend to incorporate many of the insights gained across the literature while developing our own unique contributions to this domain. Our research is primarily divided into five research areas: (1) HRL training framework and agent architecture for combat units; (2) multi-model framework for agent decision-making; (3) dimension-invariant observation abstractions of the state space; (4) intrinsic rewards engineering for our HRL framework; and (5) implementation of this framework into a higher-fidelity combat simulation. This paper focuses only on the first three research areas.

**HRL Training Framework and Agent Architecture**

We first develop an HRL training framework that will allow us to scale RL to larger and more complex scenarios by expanding upon and incorporating ideas derived from the investigations of Dayan & Hinton, (Dayan & Hinton, 1992), Vezhnevets et al. (Vezhnevets et al., 2017), Levy (Levy et al., 2019), Pope et al. (Pope et al., 2021), Wang et al. (Wang et al., 2021), Rood (Rood, 2022), and Li et al. (S. Li et al., 2022). In support of this framework, we also develop a new agent architecture consisting of an agent hierarchy and a decision hierarchy—with each individual agent being a multi-model agent.

Depicted in Figure 3, each level in the *Agent Hierarchy* primarily exerts control over a different number of subordinate agents, with the lowest level being control of a single entity. For illustration purposes, we have named these levels: *Commander*, *Manager*, and *Operator*. However, one can think of this hierarchy as anywhere from 1 to *n* levels deep, where the lowest level is 1 and the highest level is *n*. Simple tasks with few units may require only two levels, whereas complex tasks involving multiple interacting units could benefit from three or more levels of hierarchy. Because our research intends to examine more complex scenarios, we anticipate needing at least three levels of hierarchy.

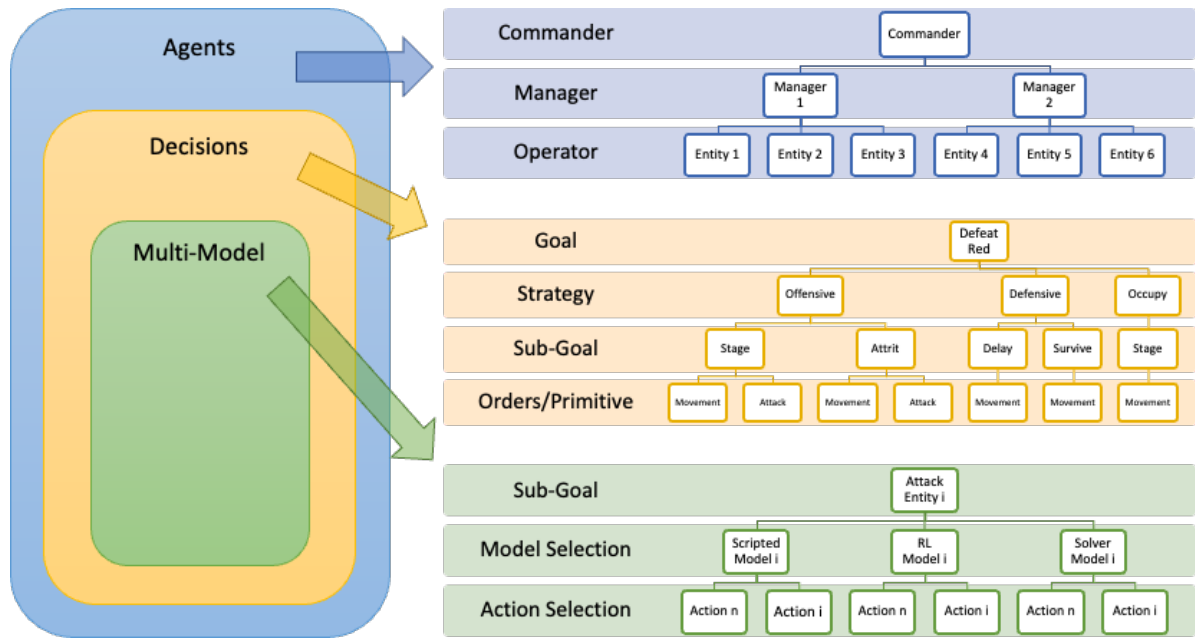

**Figure 3. Example HRL Agent and Decision Decomposition**

Within this hierarchical framework, we also develop a hierarchy of decisions. Of note, although we list specific decisions in Figure 3, these are for illustration purposes only and do not necessarily depict the final breakdown of decisions. Sutton et al. initially coined the term *options* for this concept of a hierarchy of decisions (Sutton et al., 1999). *Options* are a generalization of *actions*, which Sutton et al. formally use only for primitive choices. Previous terms have included *macro-actions*, *behaviors*, *abstract actions*, and *subcontrollers*. In representing this concept in the hierarchy, we use the term *decisions*. Whereas in a traditional RL problem an agent takes in an observation and outputs an action at regular timesteps, in an HRL problem, an agent either is given or must discover background knowledge that allows it to decompose the problem either explicitly or implicitly (Sammut & Webb, 2010). The agent then exploits this knowledge to solve the problem more efficiently by training a policy that optimizes future rewards.

Having multiple levels in a hierarchy also allows each level to be trained towards different goals and at different levels of abstraction—making scaling to very complex scenarios a more tractable problem. Additionally, this hierarchical approach implicitly trains for agent coordination and cooperation since the layer above controls the aggregate-level behaviors of the layers below (Wang et al., 2021). With the exception of the bottom-most level, each level of the hierarchy can be thought of as being abstract or cognitive (i.e., they are initially high-level decisions that will ultimately inform a primitive action). Only the agent at the bottom of the hierarchy is an actual entity on the gameboard who takes a discrete or primitive action that affects the environment.

Figure 4 illustrates our HRL framework. The *Commander* level of the hierarchy takes in its own distinct abstracted observation of the state space and outputs a subgoal and termination condition to the next level. In this next level, the *Manager* takes in the *Commander's* subgoals and a different abstracted observation of the state space and outputs its own subgoals to the next level. Finally, at the bottom-most level, the *Operator* takes in the subgoals and an abstracted local observation of the state space and outputs an action for the entity to take using our multi-model agent framework.

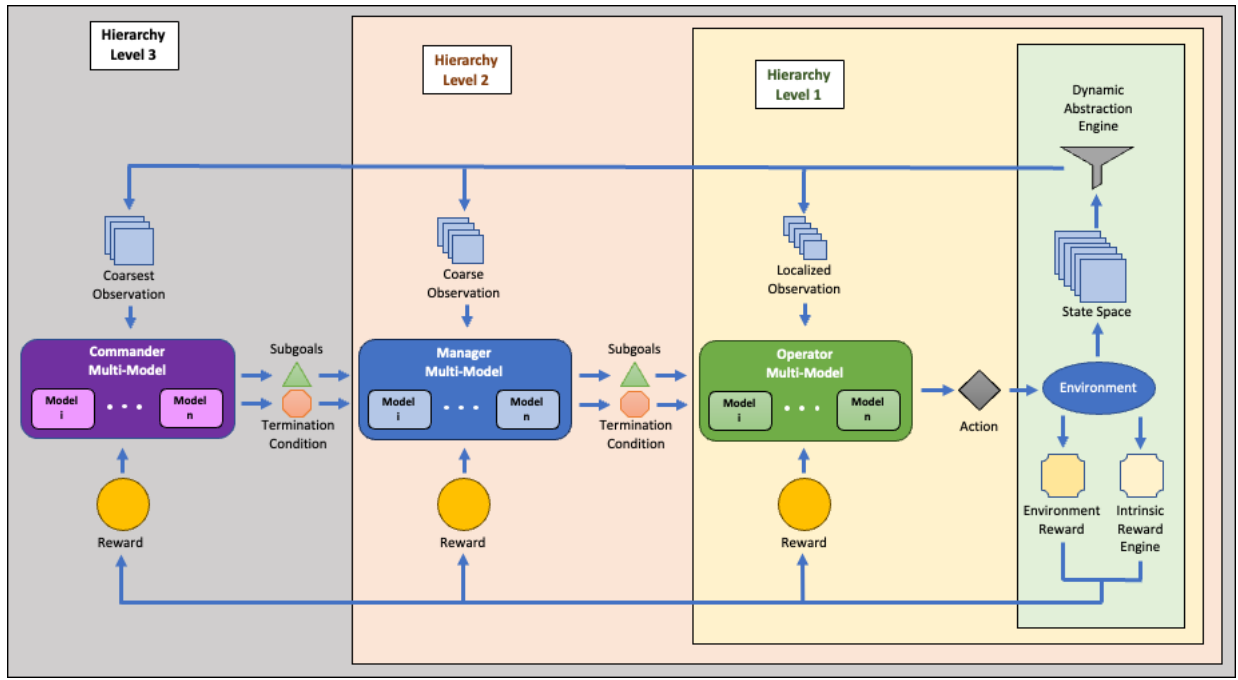

**Figure 4. Hierarchical Reinforcement Learning Framework**

**Multi-Model Agent**

To construct our *Multi-Model Agent* framework, we borrowed and employed concepts from Mixture of Experts (MoE) (Jacobs et al., 1991), "many-model thinkers" (Page, 2018), ensemble methods, and RL. Even though we borrowed the idea of employing various expert networks from the MoE and ensemble literature, we depart from the central idea these traditionally propose. While we still leverage many different models, rather than a pure divide-and-conquer approach (Jacobs et al., 1991) or a pooling of model outputs (Page, 2018), we instead differentiate between the models to determine which specific one can maximize the overall performance of the agent at each action-selection step. In other words, instead of combining model outputs, we simply use them as inputs to an evaluation function that then determines which specific behavior model should be used at each step. Whereas ensemble approaches require the modeler to account for bias or flawed models, our *Multi-Model* approach allows us to take advantage of a diverse set of models (either scripted or ML-trained) without concern for model balance or validation. This is possible because, rather than combining model predictions, we instead differentiate between the model predictions and utilize the single best policy that maximizes a specific objective.

Our *Multi-Model Framework* is depicted in Figure 5. At each action-selection step, the *multi-model* takes in an *observation* as input, and passes it to each of its *score prediction models*. Each *score prediction model* infers a *predicted game score*, which is fed into an *evaluation function*. A specific *behavior model* is then selected based on this *evaluation function*. Finally, the original *observation* is passed to the selected *behavior model*, which then produces an action.

To supply the evaluation function that selects the appropriate behavior model, we train a separate *score prediction model* for each individual behavior model in the repository. This *score prediction model* is a convolutional neural network (CNN) which infers a game score based on the current game state. This predicted game score assumes that the blue faction continues playing the game according to their respective behavior model, and the red faction continues playing according to a specific adversary behavior model. Given that Atlatl is a turn-based game and not a time-step simulation, we refer to each instance where an entity on the gameboard is prompted to take an action as an action-selection step. Although to date we have trained our *score prediction models* using supervised learning with game data, we have recently developed and are currently testing a separate version of this *score prediction model* that leverages RL instead of supervised learning—potentially enabling scalability of this approach.

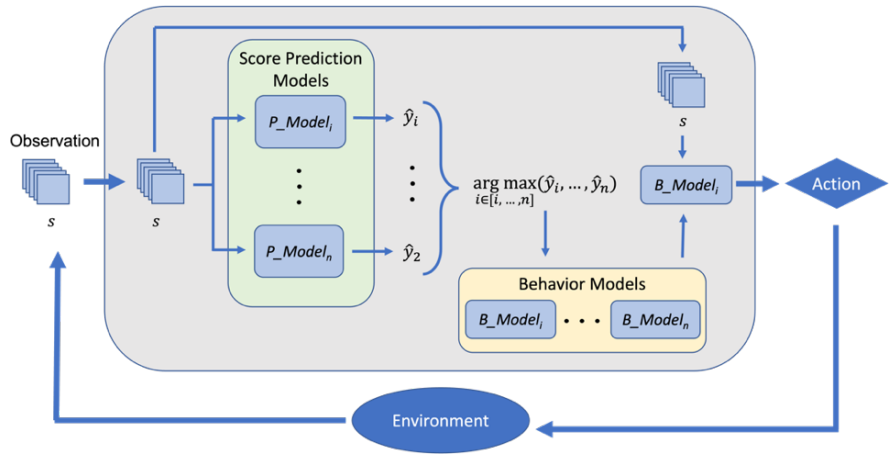

**Figure 5. Multi-Model Framework**

In our experiments that assessed the effectiveness of this multi-model approach over our traditional single-model approach (whether scripted or RL-based), we found that the multi-model approach resulted in a 62.6% improvement over our top-performing single-model. Furthermore, we found that a multi-model composed of more models significantly outperformed a multi-model composed of fewer models, even when these additional models were worse performing overall. This indicates that even though some of our individual models did not perform well in general, they were likely successful in very specific situations—a phenomena that our score prediction model seems to have captured accurately and our evaluation function used correctly to select the best model for each action-selection step.

More importantly, using this approach, rather than having to train a single model capable of performing effectively across all possible scenarios, we can instead develop or train very specialized models that can perform well in certain situations, and then automatically call on those specialized models when those specific situations (i.e., specific states in the game) are encountered. Furthermore, because our multi-model looks to differentiate between its embedded models, we can incorporate new models as needed without having to account for behavior validation, balance, or even bias—as is typically required for more traditional ensemble modeling approaches that involve pooling results.

**Observation Abstractions of the State Space**

Even using a simple environment such as Atlatl found scaling to larger scenarios resulted in poor performance (Boron, 2020; Cannon & Goericke, 2020; Rood, 2022). This is in part due to, unlike humans, RL is not sample efficient and requires large amounts of training data (Botvinick et al., 2019; Tsividis et al., 2017), which is further exacerbated by large action and observation spaces. Abel et al. noted, however, that abstractions in RL can improve sample efficiency (Abel et al., 2020), and thus potentially allow us to scale to dealing with very complex environments. Moreover, the need to learn and make use of appropriate abstract representations of the world is an essential underlying skill that will be required of any intelligent agent, whether biological or artificial (Abel, 2020).

Nevertheless, because abstractions inherently discard information—which potentially compromises the effectiveness of the decisions made based on these abstraction—we must balance the trade-off between making learning easier (or tractable) and preserving enough information to allow for optimal policy discovery (Abel, 2020). The more we abstract the state space, the more information is lost and the harder it will be to guarantee an optimal or near-optimal solution (L. Li et al., 2016). There is a tradeoff, however, in that, although coarser abstractions may result in sub-optimal actions, they do allow for better planning and value iteration (L. Li et al., 2016).

To overcome some of these tradeoff challenges of making training tractable in large state spaces while also preserving enough information to still find optimal or near-optimal solutions, we are currently developing an approach that consists of applying different levels of abstraction based on the level of the hierarchy. In *Simulation and Wargaming*,

Tolk and Laderman discuss how "[t]he task usually drives the required level of abstraction" (Turnitsa et al., 2021). Similarly, as we typically see in military planning, abstractions at the higher levels will be coarser while abstractions at the lower levels will be finer (*FM 5-0 Planning and Orders Production*, 2022; *Joint Publication 5-0 Joint Planning*, 2020; *MCWP 5-10 Marine Corps Planning Process*, 2020). Therefore, our HRL framework involves applying coarser levels of abstraction at the higher levels of decision-making while still retaining all of the local state space information for the lower levels of decision-making through local observations. We posit that this approach will allow for more effective long-term planning at the higher level, and more effective real-time local execution of those plans at the lower level.

To illustrate this concept, we first describe the general Atlatl observation space. Although constantly evolving, one of the most recent observation spaces consists of 17 channels of an *n* x *m* grid, where each entry of the grid represents one hex of an *n* x *m*-sized gameboard. This observation space is encoded as tensors. An example of what information each channel represents is depicted in Figure 6. For example, information encoded in each channel include unit on-move, unit types, terrain types, etc.

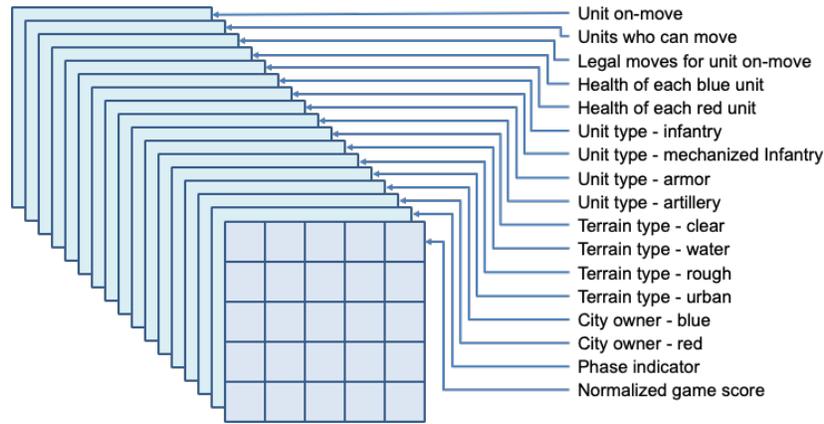

**Figure 6. Observation Space Channels**

Figure 7 shows an example state space representation of a 20 x 20 gameboard with three channels overlayed (blue forces, red forces, urban hex). We overlay the three channels in all of figures below for illustration purposes only; however, within Atlatl, these would be represented as 3 independent channels (out of 17 channels total) as depicted in Figure 6.

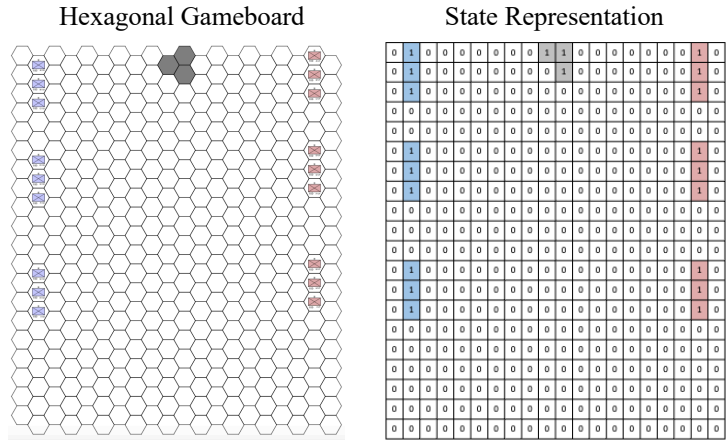

**Figure 7. State Space Representation Example of a 20 x 20 Gameboard**

An example of how we can abstract the space in a dimension-invariant way that can be used for any map size is shown in Figure 8. On the left side is a coarse observation of the gameboard (to be used by our Commander agent), where the 20 x 20 grid has been divided up into a 5 x 5 grid and the values summed up per the overlayed coarse grid.

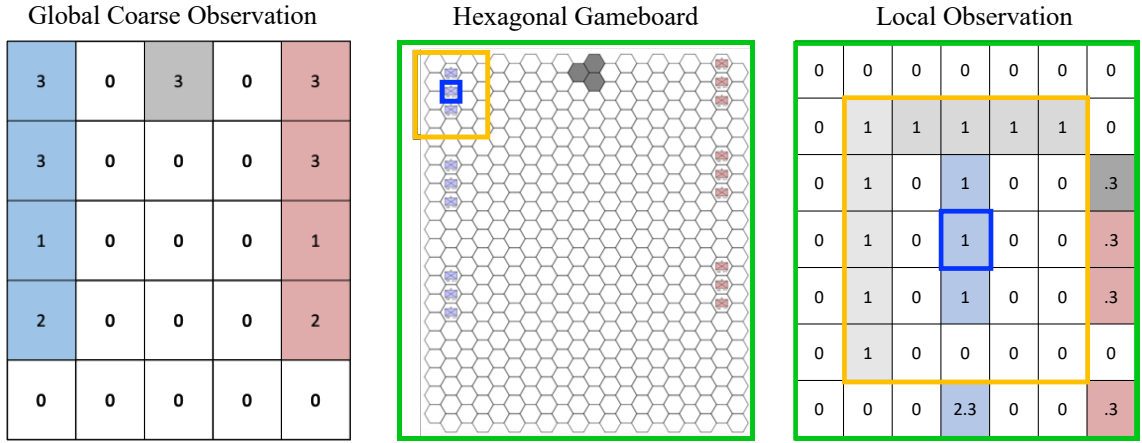

**Figure 8. Coarse Abstraction Observation and Localized Observation**

On the right side in Figure 8 is a localized observation (to be used by our Operator agent) using a 7 x 7 grid where the entity on-move is now centered in the observation space (depicted by the blue outline) and the surrounding 2 grids are geographically to scale (depicted within the orange outline). For our outmost layer of the observation, we use a piecewise linear spatial decay function that calculates weights based on how far away the item of interest is from the unit on-move, as depicted in Figure 9. These weights are then multiplied by the values within each respective layer, summed up by radial, and then inserted into the outermost layer of the local 7 x 7 grid. We only show two observation abstractions above, however, we intend to include at least three different levels of observation

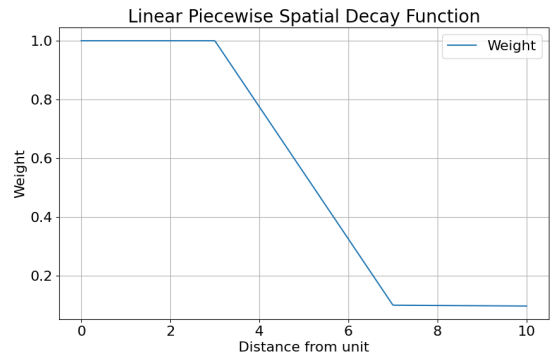

**Figure 9. Weights for the Piecewise Linear Spatial Decay Function**

abstractions (not shown is an abstracted 7 x 7 global observation to be used by our Manager agent).

Although experiments are still ongoing, initial results are promising. For example, for our initial experiment we use a 10 x 10 gameboard with 2 to 4 agents per side, 1 city, and 30 phases. We start with training both a global-observation and a local-observation agent (using a local 7 x 7 observation space) for a total 2 million steps on a single scenario (i.e., the same initial starting conditions). Our preliminary results from evaluating our baseline scripted agent and our RL-trained agents across 1,000 games is shown in Table 1.

**Table 1. Preliminary Experiment Results**

|  | **Baseline Scripted Agent** | **RL-Global Observation Agent** | **RL-Local 7x7 Observation Agent** |
|---|---|---|---|
| **Mean Score** | 813.750 | 1095.601 | 1070.512 |
| **Standard Deviation** | 6.253 | 62.972 | 24.992 |

For this specific experiment set-up, we see that both RL-trained agents outperformed the baseline scripted agent, with the global-observation agent performing best. This result is not unexpected. We do anticipate that due to the loss of fidelity in information by the local-observation agent, the global-observation agent will likely outperform the local-observation agent given an unlimited training budget (i.e., convergence to an optimal policy). Our goal, however, is to determine whether an agent trained on a local observation space can outperform a global-observation agent in complex scenarios with a limited training budget (i.e., where convergence to an optimal policy cannot be guaranteed).

Although this is a promising start to show that an agent may be able to learn to select the appropriate behaviors by simply using a local observation space, it remains to be seen if this ability to learn from this reduced representation of the state space is generalizable across any starting condition. Thus, we are currently conducting an experiment with the same gameboard set-up but using an unlimited number of possible scenarios (i.e., completely random starting states) and increasing our training budget to 5 million steps. In the preliminary training results of this experiment, as depicted in Figure 10, we find that the global-observation agent learns very slowly while the three local-observation agents we tested are able to learn relatively quickly. The local observation used that yields the best performance is the local 7 x 7 observation using the piecewise linear spatial decay. However, we are currently in the process of testing the performance of these agents against our baseline scripted agent.

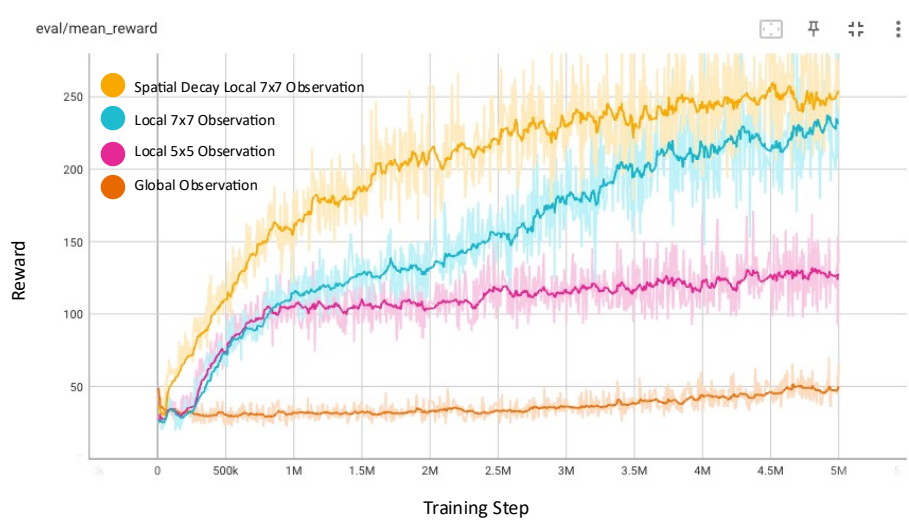

**Figure 10. Learning Rate of Local Observation Agents Versus a 10 x 10 Global Observation Agent**

We must note that although this graph shows learning is occurring, we acknowledge that it may not lead to significantly better performance in practice over our baseline scripted agent. Nevertheless, because our current engineered RL rewards for this experiment are based almost entirely on the game score, there exists a correlation in this case between rewards and expected performance. Moreover, what is also promising in this local-observation approach is that, because we are using a dimension-invariant observation space, we can use an agent that is trained on a 10 x 10 gameboard (or any sized gameboard) on much larger gameboards and achieve relatively similar performance—assuming similarities in scenarios. Furthermore, because we ultimately intend to combine these local-observation agents with our Multi-Model approach, it now becomes possible to more quickly train local-observation agents on specialized scenarios using a smaller training budget than that which would be required of global-observation agents.

**ONGOING AND FUTURE WORK**

Our next steps involve continuing to develop and identify the best types of observation abstractions for each level of the hierarchy. As defined by Abel, "a *state abstraction* determines which changes to the environment count as substantive" (Abel, 2020). Thus, our goal is to continue experimenting and determine which levels of abstraction are appropriate based on the decisions to be made at each level of the hierarchy—from the operational level down to the tactical level.

Once we have selected the observation abstractions that lead to an acceptable tradeoff between performance and training based on the respective level of the hierarchy being computed, we will determine the inputs to the agents on each level of the hierarchy as well as their respective action spaces. We anticipate that inputs to the agent at the Commander level will consist of the ultimate objective and a coarse observation. For example, this objective may be to maximize game score, defeat all red forces, preserve all blue forces, or control all urban areas. Example outputs from the Commander level, that would serve as inputs to the Manager level, may include posture (e.g., offensive, defensive, deterrent, preemptive) and force assignment (e.g., which group of units should be assigned to each area or task). With these inputs (i.e., sub-tasks) and the appropriate observation abstraction, the Manager's action space will likely consist of sub-tasks such as objective areas that units under its control should move to, as well as their specific tasks once they reach these locations. These Managers' sub-tasks would then become the input to the final level of the hierarchy where each Operator-level unit would use its Multi-Model framework to output a specific action to take.

Following this, we intend to develop the *Intrinsic Reward Engine* portion of our HRL framework. This reward engine will allow us to incentivize agents during training to accomplish specific goals or tasks per phase regardless of how it may affect the overall score of the game in the long run. With this piece, we will then begin to train the HRL framework beginning at the lowest level and working towards the highest level. We will only train one level of the hierarchy at a time while freezing all other levels to account for the known RL non-stationarity problem (Henderson et al., 2019) that results from multiple levels trying to learn different policies simultaneously.

Along the way, we will also continue to refine each of the components of our HRL framework. For example, we will implement our RL-based Multi-Model framework and further break up the score predictions by phase (instead of by game) to allow for different goals to be pursued per phase. Additionally, we will develop, train, and incorporate more specialized models.

Overall, our research to date shows promise in this approach to scale the application of AI to very complex domains such as that of combat modeling and simulation in support of wargaming. Our Multi-Model framework has already allowed us to drastically improve the performance of our agents well beyond our current state-of-the-art scripted or RL-trained agents alone within our combat simulation. Our local observation abstraction of the state space shows potential in that, given a reasonable training budget, it now allows agents to learn in much bigger gameboards than have previously been possible within Atlatl. We anticipate that incorporating these developments into our overall HRL framework will further improve the performance of our agents and potentially allow us to scale to virtually any size scenario.